# Color-Stripe Structured Light Robust to Surface Color and Discontinuity

Kwang Hee Lee    Changsoo Je    Sang Wook Lee

Dept. of Media Technology
Sogang University
Shinsu-dong 1, Mapo-gu, Seoul 121-742, Korea
{Khecr, Vision, Slee}@sogang.ac.kr

**Abstract.** Multiple color stripes have been employed for structured light-based rapid range imaging to increase the number of uniquely identifiable stripes. The use of multiple color stripes poses two problems: (1) object surface color may disturb the stripe color and (2) the number of adjacent stripes required for identifying a stripe may not be maintained near surface discontinuities such as occluding boundaries. In this paper, we present methods to alleviate those problems. Log-gradient filters are employed to reduce the influence of object colors, and color stripes in two and three directions are used to increase the chance of identifying correct stripes near surface discontinuities. Experimental results demonstrate the effectiveness of our methods.

## 1    Introduction

Structured light-based ranging is an accurate yet simple technique for acquiring depth image, and thus has been widely investigated [3-5, 7-15, 18]. There has been a variety of light patterns developed for rapid range imaging and the range resolution achievable in a single video frame has recently been increased sufficiently so that real-time shape capture became a practical reality. [12, 14, 18].

Structured-light methods suitable for realtime ranging uses their specific codifications based on color assignment [3, 5, 8, 11, 12, 14, 15, 18] and *spatially windowed uniqueness* (SWU) [3, 8, 12, 14, 18] for increasing the number of unique subpatterns in a single projection pattern. This SWU and the use of color have made it possible to design color stripe patterns for high-resolution range imaging in a single video frame. On the other hand, the main disadvantages of a single-frame color stripe pattern compared to sequential BW (black-and-white) stripes are that color stripe identification is affected by object surface color and it is often impossible near surface discontinuities such as occluding boundaries. In general, in a multiple-stripe color pattern, the uniqueness of a spatially windowed subpattern becomes higher as the number of stripes in the subpattern increases. To guarantee sufficient SWU for rapid imaging, a subpattern consists of several adjacent stripe colors. However, some of those adjacent stripes are often unavailable to identify a stripe color of interest near surface discontinuities. To the best of our knowledge, little research has been carried out explicitly to alleviate these problems.

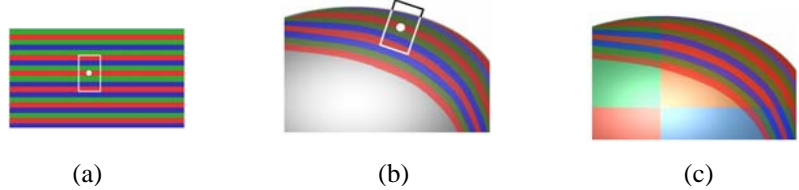

(a)  (b)  (c)

**Fig. 1.** (a) A multiple-color stripe pattern and a unique subpattern with 7-stripe wide spatial window, (b) a subpattern window in projected stripes, and (c) projected stripes on a colored object

In this paper, we present structured-light methods with range estimation robust to object surface colors and discontinuities. We show that the logarithmic gradient (log-gradient) [1, 2] can be used for decreasing the influence of surface color reflectance in structured-light range imaging. Our data processing algorithm incorporates the Fourier transform-based filtering that has been used for deriving intrinsic images [6, 16, 17]. By applying the algorithm to a captured image, the object reflectance in the image is discounted to some degree, and the stripe colors of structured light are detected with higher accuracy. In addition, we develop a method of applying stripe patterns in two or three directions simultaneously, decoupling the patterns and estimating depth images independently. This substantially improves the chance of estimating correct depths near surface discontinuities.

The rest of this paper is organized as follows. Section 2 presents the problem statement, and Section 3 presents a filtering processing method for decreasing the influence of surface colors in color stripe identification. A method of applying more than one pattern simultaneously for improving depth estimation near surface discontinuities is described in Section 4. Experimental results are shown in Section 5 and Section 6 concludes this paper.

## 2   Problem Statement

Sequential projections of BW patterns are in general inappropriate for rapid structure-light depth imaging and the use of colors has been investigated to reduce the number of pattern projections required for a given range resolution. If a stripe by its color alone in a color stripe pattern, global uniqueness is hard to achieve due to the repeated appearance of the color among $M$ stripes. The use of more colors than three RGB can compromise their distinctiveness due to the reduced color distances between the stripes. In the SWU scheme, colors are sequentially encoded to stripes according to permutations or de Bruijn sequence. A stripe color centered a spatially windowed subpattern can be identified since the combination of the colors in the window is unique globally or at least semi-globally [3, 8, 12, 14, 18]. Figure 1 (a) illustrates this uniqueness of a seven-stripe subpattern. The larger the number of stripes is in the subpattern, the wider the window is and the more unique the subpattern is.

A wide subpattern has an obvious disadvantage near a surface discontinuity. When the directions of an occluding boundary and stripes are similar, some stripes in the subpattern around a color stripe may be occluded and thus the color stripe is unidentifiable as depicted in Figure 1 (b). This results in a failure in depth estimation near surface discontinuities. The failure also occurs when strong surface colors alter stripe colors significantly as shown in Figure 1 (c). The goal of the research presented in this paper is to develop algorithms for reducing the influence of object colors and discontinuities using log-gradient filters and combined projection patterns in multiple directions.

## 3  Log-Gradient Processing for Surface Color Reduction

As mentioned earlier, surface color hinders detection of stripes. Angelopoulou *et al.* proposed a spectral log-gradient method [1], and Berwick and Lee used spectral and spatial log-gradients to discount illumination colors [2]. In the research for deriving intrinsic image, several filtering schemes have been proposed [6, 16, 17]. Based on those spectral and spatial operators and filtering algorithms, we develop a method for discounting object colors across the color stripe directions.

Let $p_d$ be the result signal of an original signal $p$ processed by a filter $f_d$:

$$p_d \equiv f_d * p. \tag{1}$$

Then, $p$ can be estimated as follows [6]:

$$\hat{p} = F^{-1}\left( \frac{F(\sum_{d=x,y} f_d^r * \hat{p}_d)}{F(\sum_{d=x,y} f_d^r * f_d)} \right), \tag{2}$$

where $f_d^r$ is the reversed filter of $f_d$: $f_d(x,y) = f_d^r(-x,-y)$, and $F$ and $F^{-1}$ denote the Fourier transform and its inverse transform, respectively.

The objective of our processing algorithm is to regard the color variation across the stripes as illumination change (i.e., projection color change) and remove the surface reflectance as much as possible. We use log-gradient operators for this purpose. The reflected light intensity is given by the equation:

$$I(\lambda) = S(\lambda)L(\lambda), \tag{3}$$

where $S$, $L$ and $\lambda$ is the surface reflectance, illumination and wavelength, respectively. Applying log-gradient with respect to the way of stripe transition to the equation 3,

$$\partial_x(\ln I) = \partial_x(\ln S) + \partial_x(\ln L) \simeq \partial_x(\ln L), \tag{4}$$

where it is assumed the spatial change of surface reflectance is much slower than that of illumination (through the stripe transition) and can be ignored. In equation 4, we can see spatially static surface reflectance can be easily removed. Since the stripe direction on the surface spatially varies in a scene image due to the surface geometry and triangulation, log-gradient has to be applied w.r.t. the corresponding direction to each infinitesimal area for removing the surface reflectance component as much as possible. We do this process approximately by differentiating the images rotated from the original image with 17 angles sampled at the step of 10 degrees. In order to operate this process based on a filtering scheme in intrinsic imaging we apply a gradient filter w.r.t. $x$, $D_x$ to the logarithm of each rotated image:

$$i_x^\varphi \equiv D_x * \ln\left(R_\varphi(I)\right), \tag{5}$$

where $R_\varphi$ denotes the rotation operator by an angle $\varphi$. For obtaining depth data it is proper to restore the illumination component of single-directional transitions. Therefore, in our application the restoration does not include the gradient w.r.t. $y$, $i_y^\varphi$ contrary to equation 2:

$$\tilde{i}^\varphi \equiv F^{-1}\left(\frac{F(D_x^r * i_x^\varphi)}{F(\sum_{d=x,y} D_d^r * D_d)}\right). \tag{6}$$

Taking the exponential to the restoration and rotating back each image by $-\varphi$, the restoration of each rotation is obtained:

$$\tilde{I}^\varphi \equiv R_{-\varphi}\left(\exp(\tilde{i}^\varphi)\right). \tag{7}$$

Finally, the illumination-restored image can be constructed by $\tilde{I}^{\Phi(x,y)}$ where $\Phi(x,y)$ is the rotation angle which maximizes the gradient magnitude:

$$\Phi \equiv \arg\max_\varphi \sqrt{(\partial_x \tilde{I}^\varphi)^2 + (\partial_y \tilde{I}^\varphi)^2}. \tag{8}$$

With the illumination-restored image, stripe segmentation is expected to be much less insensitive to object surface colors and to result in more accurate depth results.

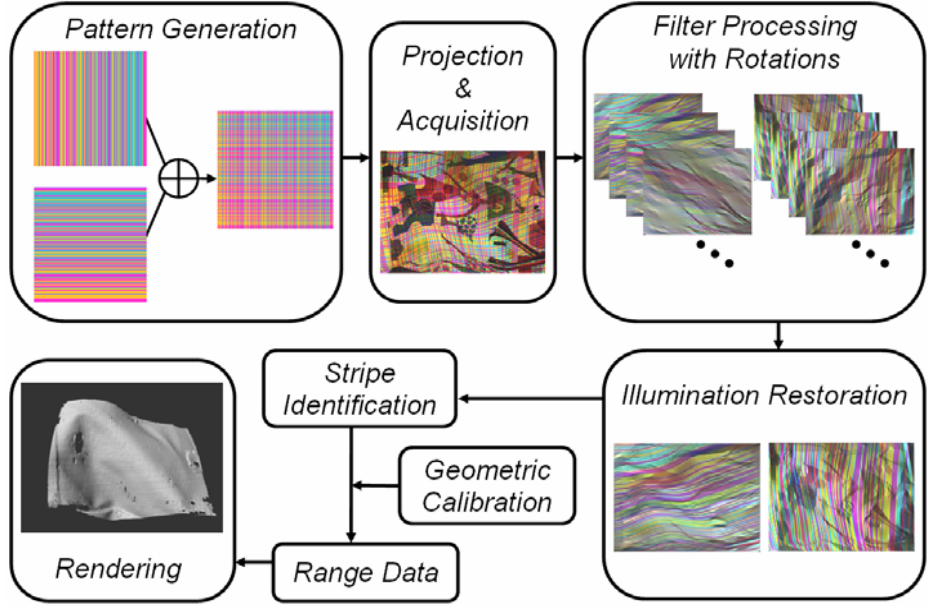

**Fig. 2.** The process of 3D modeling based on the proposed method with a combination of two stripe-patterns. Horizontal and vertical color stripe-patterns encoded based on SWU, are combined to a single combination pattern, and it is projected onto the leather surface of a color-textured bag. The scene is captured by a camera, and the scene image is filter-processed with rotations to estimate the two illumination-restored images. Stripes are identified independently in each restoration image, and ranges are obtained by geometric calibration of the projector and camera. Finally the ranges are merged into one, and it is meshed and rendered.

## 4 Patterns Combination for Adapting to Surface Discontinuities

As asserted in section 2, SWU based stripe identification can lose the correct coordinate of illumination near the surface discontinuity. In a scene where dominant discontinuities exist through the horizontal way, projection of vertical stripes can easily escape the problem of SWU based identification. By combining two sets of stripes with different directions, a single projection pattern containing multidirectional stripes can be generated, and projected into a scene. As a result, it can be noted that for any of single-directional discontinuities there can be a sufficient number of connected stripes with the directions distant to those of discontinuities, and thus the stripes can be correctly identified for most of single-directional discontinuities. In this case, we can find the two colors of the illuminated stripes, $\tilde{I}^{\Phi 1}$ and $\tilde{I}^{\Phi 2}$ by estimating $\Phi 1$ and $\Phi 2$ for each pixel. We estimate $\Phi 1$ and $\Phi 2$ by a similar way as in equation 8, by finding the maximum in the $\pi/2$ period centering at the globally dominant direction ($m_{\Phi_1}$ and $m_{\Phi_2}$) of each stripe-pattern. In an appropriate case, stripe patterns with three different directions can be combined into a single pattern to be projected into the scene, and $\Phi 1$, $\Phi 2$ and $\Phi 3$ can be found in each smaller range ($\pi/3$ interval) with the center $m_{\Phi 1}$, $m_{\Phi 2}$, and $m_{\Phi 3}$. Figure 2 illustrates the process of 3D modeling based on the proposed method with a combination of two stripe-patterns. The process is mostly similar to those for a single stripe-pattern and three intrinsic patterns.

## 5  Experimental Results

We have made many experiments to validate the proposed techniques. The experimental setup consists of a Sony XC-003 3CCD VGA camera, an Epson color LCD XGA projector and a cubic calibration-object. A permutation-based color-stripe pattern in [12] is used to generate the three kinds of patterns: single-direction, double-direction and triple-direction stripe-patterns (see Figure 3).

We used the Sobel operator as the gradient filter, and the stripes are segmented by hue thresholding of restored illumination colors normalized by neighbor colors. Each segmented stripe is identified according to the sequence of neighboring consecutive stripes. From the image coordinates and stripe identities, 3D world coordinates are estimated by the geometric calibration [13] which gives the extrinsic and intrinsic parameters of the projector and camera. When the double-direction or triple-direction is used, more than one range image are obtained, and are merged into one by removing erroneous points and by averaging multiple reliable points. Some results are meshed and rendered.

Figure 4 shows the experimental results of a crumpled paper which consists of squares of 4 colors (cyan, blue, green and red, clockwise): (a) input image, (b) illumination-restored image by log-gradient processing without rotations (c) illumination-restored image by proposed processing, (d) range result by direct estimation (e) range estimation from (b), and (f) range estimation from (c). Figure 5 compares the stripe segmentation results by direct estimation and by proposed processing.

Figure 6 shows the results of the leather surface of a color-textured bag: (a) input image, (b) illumination-restored image by proposed processing, (c) scene under white illumination, (d) rendered result by direct estimation, (e) result by proposed method, and (f) a filtered image with a single rotation.

Figure 7 depicts the results of the bag: (a) input image, (b and c) the two illumination-restored images, (e and f) rendered results from (b and c), and (d) merged result from (b and c).

Figure 8 depicts the results of hands: (a) input image, (b and c) the two illumination-restored images, (e and f) rendered results from (b and c), and (d) merged result from (b and c).

Figure 9 depicts the results of results of a cow: (a) input image of a scene with double-directional pattern, (b and c) the two illumination-restored images, (e and f) rendered results from (b and c), and (d) merged result from (b and c).

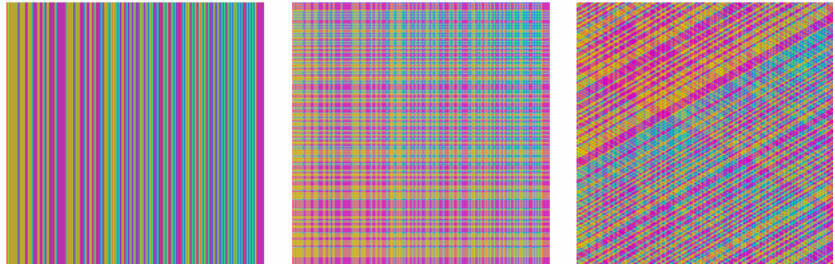

**Fig. 3.** The three kinds of stripe-patterns: (Left) single-direction, (Middle) double-direction, and (Right) triple-direction.

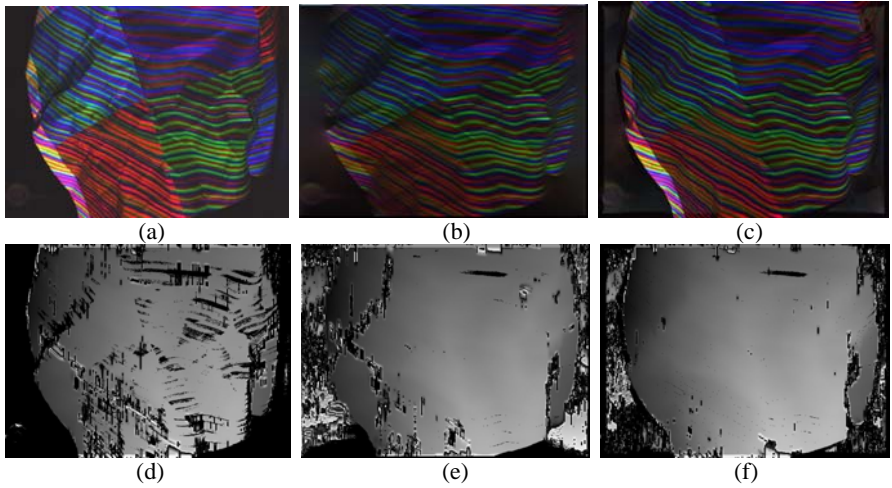

**Fig. 4.** The results of a crumpled paper which consists of squares of 4 colors (cyan, blue, green and red, clockwise): (a) input image, (b) illumination-restored image by log-gradient processing without rotations (c) illumination-restored image by proposed processing, (d) range result by direct estimation (e) range estimation from (b), and (f) range estimation from (c).

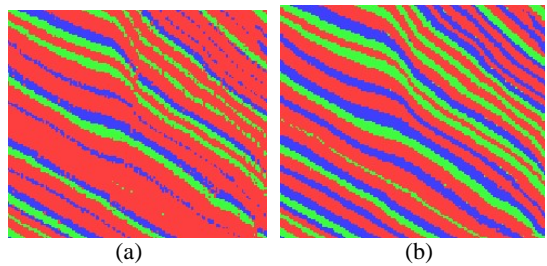

**Fig. 5.** The stripe segmentation results by (a) direct estimation and by (b) proposed processing.

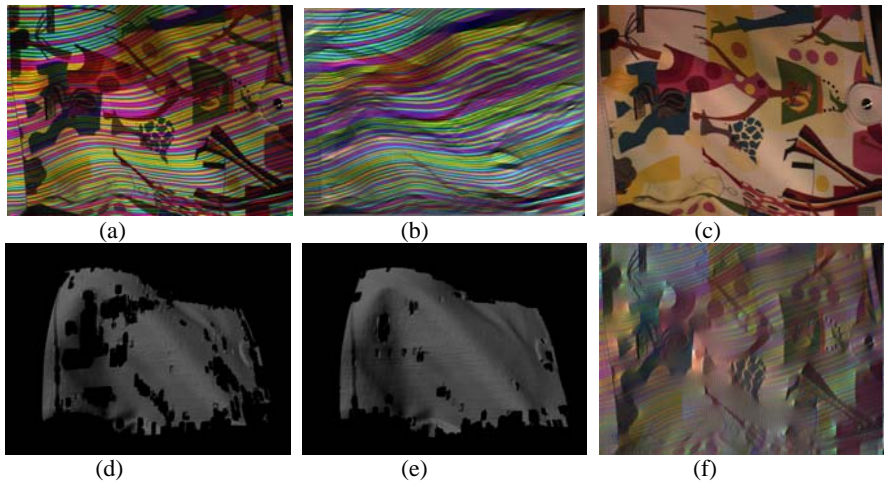

**Fig. 6.** The results of the leather surface of a color-textured bag: (a) input image, (b) illumination-restored image by proposed processing, (c) scene under white illumination, (d) rendered result by direct estimation, (e) result by proposed method, and (f) a filtered image with a single rotation.

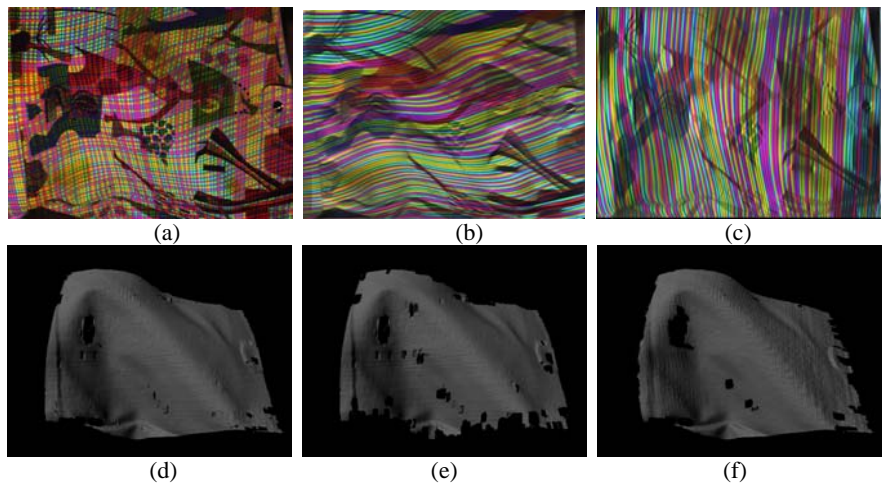

**Fig. 7.** The results of the bag: (a) input image, (b and c) the two illumination-restored images, (e and f) rendered results from (b and c), and (d) merged result from (b and c).

**Acknowledgments.** This work was supported by the Korea Science and Engineering Foundation (KOSEF) grant funded by the Korea government (MOST) (No. R01-2006-000-11374-0) and Seoul R&D Program.

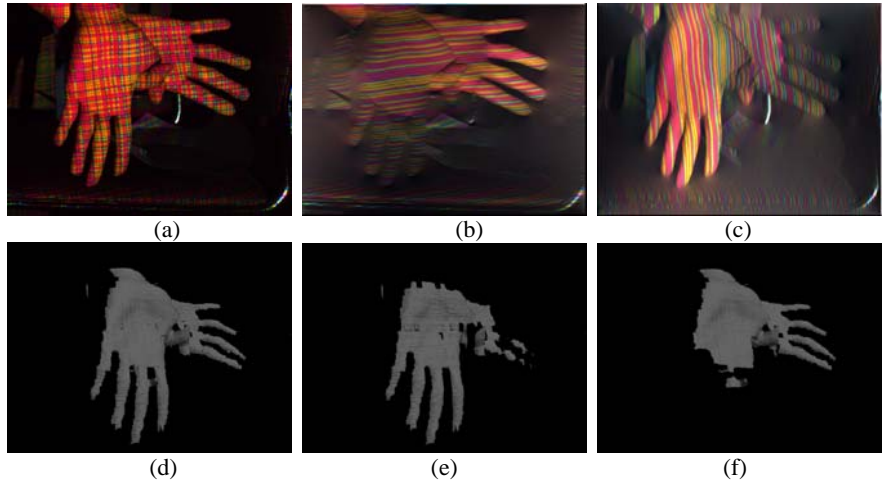

**Fig. 8.** The results of hands: (a) input image, (b and c) the two illumination-restored images, (e and f) rendered results from (b and c), and (d) merged result from (b and c).

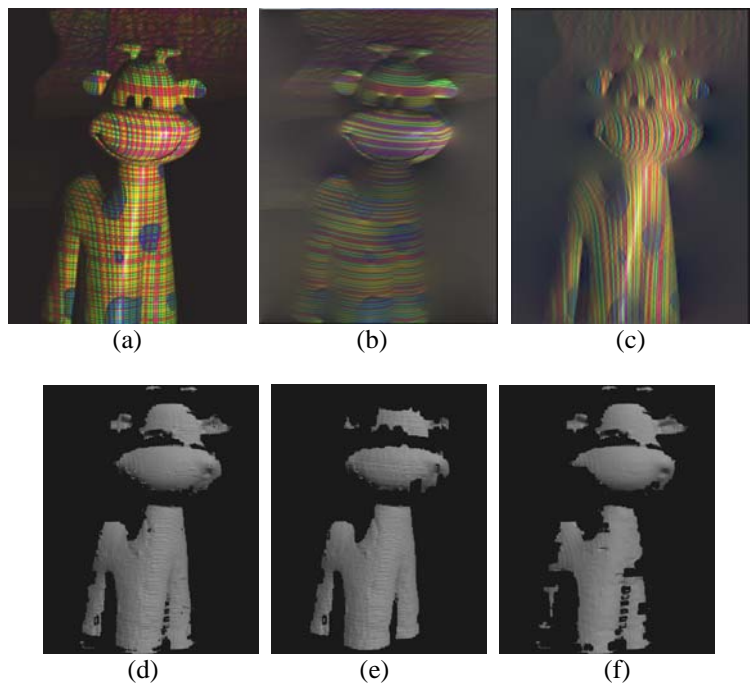

**Fig. 9.** The results of a cow: (a) input image of a scene with double-directional pattern, (b and c) the two illumination-restored images, (e and f) rendered results from (b and c), and (d) merged result from (b and c).

## 6   Conclusion

In this paper, we demonstrated a filtering method for improving the estimation of stripe colors, and proposed a strategy of combining two or three stripe-patterns for increasing the probability of correct stripe identification. Through the experiments, we have shown that the method makes the range acquisition more insensitive to the scene characteristics.